\documentclass[journal]{IEEEtran}

\ifCLASSINFOpdf
\else
   \usepackage[dvips]{graphicx}
\fi
\usepackage{url}
\usepackage{multirow}

\usepackage{graphicx}
\usepackage{bm}
\usepackage[table,xcdraw]{xcolor}
\begin{document}

\title{Deep Learning-based Inertial Odometry for Pedestrian Tracking using Attention Mechanism and Res2Net Module}

\author{Boxuan Chen, Ruifeng Zhang, Shaochu Wang, Liqiang Zhang*, Yu Liu

\thanks{Boxuan Chen, Ruifeng Zhang, Liqiang Zhang* and Yu Liu are with School of Microelectronics, Tianjin University, Tianjin 300072, China.}
\thanks{Shaochu Wang is with the Tianjin Institute of Surveying and Mapping Co. Ltd, Tianjin 300160, China. (Corresponding authors: Liqiang Zhang, E-mail: zhangliqiang@tju.edu.cn).}
}

\markboth{}
{Shell \MakeLowercase{\textit{et al.}}: Bare Demo of IEEEtran.cls for IEEE Journals}
\maketitle

\begin{abstract}
Pedestrian dead reckoning is a challenging task due to the low-cost inertial sensor error accumulation. Recent research has shown that deep learning methods can achieve impressive performance in handling this issue. In this letter, we propose inertial odometry using a deep learning-based velocity estimation method. The deep neural network based on Res2Net modules and two convolutional block attention modules is leveraged to restore the potential connection between the horizontal velocity vector and raw inertial data from a smartphone. Our network is trained using only fifty percent of the public inertial odometry dataset (RoNIN) data. Then, it is validated on the RoNIN testing dataset and another public inertial odometry dataset (OXIOD). Compared with the traditional step-length and heading system-based algorithm, our approach decreases the absolute translation error (ATE) by 76$\%$-86$\%$. In addition, compared with the state-of-the-art deep learning method (RoNIN), our method improves its ATE by 6$\%$-31.4$\%$.
\end{abstract}


\IEEEpeerreviewmaketitle

\section{Introduction}

\IEEEPARstart{W}{hether} in consumer-level services or professional service scenarios, there are fast and accurate indoor positioning requirements, including shopping guides in shopping malls, parking lot car search, human-computer interaction, and crowd monitoring. Many methods rely on communication networks, for example, a WLAN-based approach is used to statistically analyze the signal strength fingerprint of buildings \cite{Yang2012LocatingIF,Au2013IndoorTA}. However, all these methods rely on external signals, and the results are not stable. While pedestrian dead reckoning (PDR) technology \cite{Harle2013ASO} only relies on the inertial measurement unit (IMU), its superior flexibility and portability make it attract increasing attention as a practical method. 




The PDR approach commonly includes the inertial navigation system (INS) and step-length and heading system (SHS) methods. INS \cite{Foxlin2005PedestrianTW} is based on Newtonian mechanics to determine the position at each moment. However, its errors accumulate very rapidly. By contrast, the SHS algorithm \cite{Tian2016AMD} measures the length and heading of each step and then updates the current position. Its error accumulation characteristic is better than that of INS. To mitigate the position drift caused by error accumulation, many works integrate IMUs with other sensors to improve accuracy, such as the visual-inertial odometer \cite{Leutenegger2015KeyframebasedVO}, the WIFI-integrated inertial odometer, etc. Some studies have reduced drift using step length without the help of other sensors, based on prior knowledge of human walking. One of the methods is zero velocity updates (ZUPT) \cite{Foxlin2005PedestrianTW}, but ZUPT requires a foot-mounted sensor to achieve high accuracy. Another category is step counting \cite{Brajdic2013WalkDA}, which does not require the sensor to be linked to the foot. But it needs many parameters debugging and is not applicable in daily life.

Deep learning-based methods do not require manual parameters during testing and transform the problem of inertial navigation into a continuous time-series learning task. RIDI \cite{Yan2018RIDIRI} first combines inertial navigation with machine learning to estimate position by learning linear acceleration and angular velocity regression velocity vectors, then the linear acceleration is corrected by linear least squares. IONet \cite{Chen2018IONetLT} turns the inertial navigation problem into a time-series deep learning problem for the first time. The position is determined directly by the speed and direction of the network regression without using the traditional integration method, and the uncertainty is regressed in their subsequent work \cite{Chen2021DeepNN}. Yan \emph{et al.} \cite{Yan2020RoNINRN} finish an extensive and pose-rich collection of accurate trajectories under natural human motion and propose three new state-of-the-art models (RoNINs) that ultimately regress the position. The present work demonstrates that the deep learning model can infer motion trajectories from original IMU data. IONet, RoNIN, and other networks get better results in inferring trajectories directly, suggesting the possibility of deep learning localization. We continue to study how to improve its accuracy based on the previous work. This paper constructs a deep learning-based inertial odometry for pedestrian tracking, and our contributions can be summarized as the following:

\begin{itemize}
\item We build a novel deep learning-based inertial odometry for pedestrian tracking with a velocity regression neural network and integrate velocities to generate positions.

\item Two convolutional block attention modules (CBAMs) and several Res2Net modules are integrated into the velocity regression network to enhance feature extraction and represent features with fine granularity to make the network more focused on valuable information and improve the quality of the output.
\end{itemize}

 


\section{Method}
We design a velocity estimation network and then integrate velocities to generate positions.
\subsection{Velocity Estimation Network}
\subsubsection{Pipeline of Our Proposed Model}
We propose a neural network to learn the relationship between raw IMU data and horizontal velocity vector. Our network uses a 1D version of ResNet50 architecture proposed in \cite{He2016DeepRL} modified with the Res2Net module and CBAM. The entire model structure is illustrated in Fig. 1.

The network input dimensions are $N\times6$ continuous IMU raw data, and $N$ is a time window reflecting the temporal characteristics of the motion. The output of the network is a two-dimensional vector: the velocity $\upsilon_x$ in the horizontal $x$ direction and the velocity $\upsilon_y$ in the horizontal $y$ direction. Effectively, the neural network acts as a function $g_\theta$, which maps the raw IMU data to the horizontal velocity vector over a sample window. 

\begin{equation}
    (\boldsymbol{f},\boldsymbol{w})_{N\times6}\stackrel{g_\theta}{\longrightarrow}(\upsilon_x,\upsilon_y)_{1\times2}
\end{equation}

The proposed network consists of three main structures. The first input module contains the parts from a one-dimensional convolutional layer (Conv1d) to a maximum pooling layer (MaxPool). This part aims to extract the features of specific force ($\boldsymbol{f}$) and angular rate ($\boldsymbol{w}$) and reduce the dimension of the features. The second part is the attention mechanism module (CBAM), which consists of the channel attention module (CAM) and the spatial attention module (SAM). Through the attention module, the feature expression can be improved. The third part is LayerX (X=1, 2, 3, 4), composed of several Res2Net modules. 


The feature representations learned in the input module move to Layer1 that consists of several Res2Net modules. As shown in Fig. 1, three Res2Net modules are connected in series to form Layer1. Layer2, Layer3, and Layer4 are similar in structure to Layer1. The only difference is the number of Res2Net modules. We follow the original design of ResNet50 to set the number of Res2Net modules in each Layer to 3, 4, 6, and 3. The features pass through a network of three layers in series and then move to a CBAM. Then, the processed features can be projected into the velocity vector after Layer4, a CBAM and a fully connected (FC) layer. 

The application of Conv1d is to extract time-level feature relationships from continuous IMU data. Batchnorm1d is leveraged to normalize the extracted features to eliminate the effects of data variation. We make use of the mean square error (MSE) as the loss function during training, and the MSE loss is defined as:
\begin{equation}
    \ell=\frac{1}{n}\sum_{i=1}^n\left(\Vert \upsilon_{xi}-\hat{\upsilon}_{xi}\Vert^2+\Vert \upsilon_{yi}-\hat{\upsilon}_{yi}\Vert^2\right)
\end{equation}

\begin{figure}[t]
  \centering
  \label{fig1}
  \includegraphics[width=9cm]{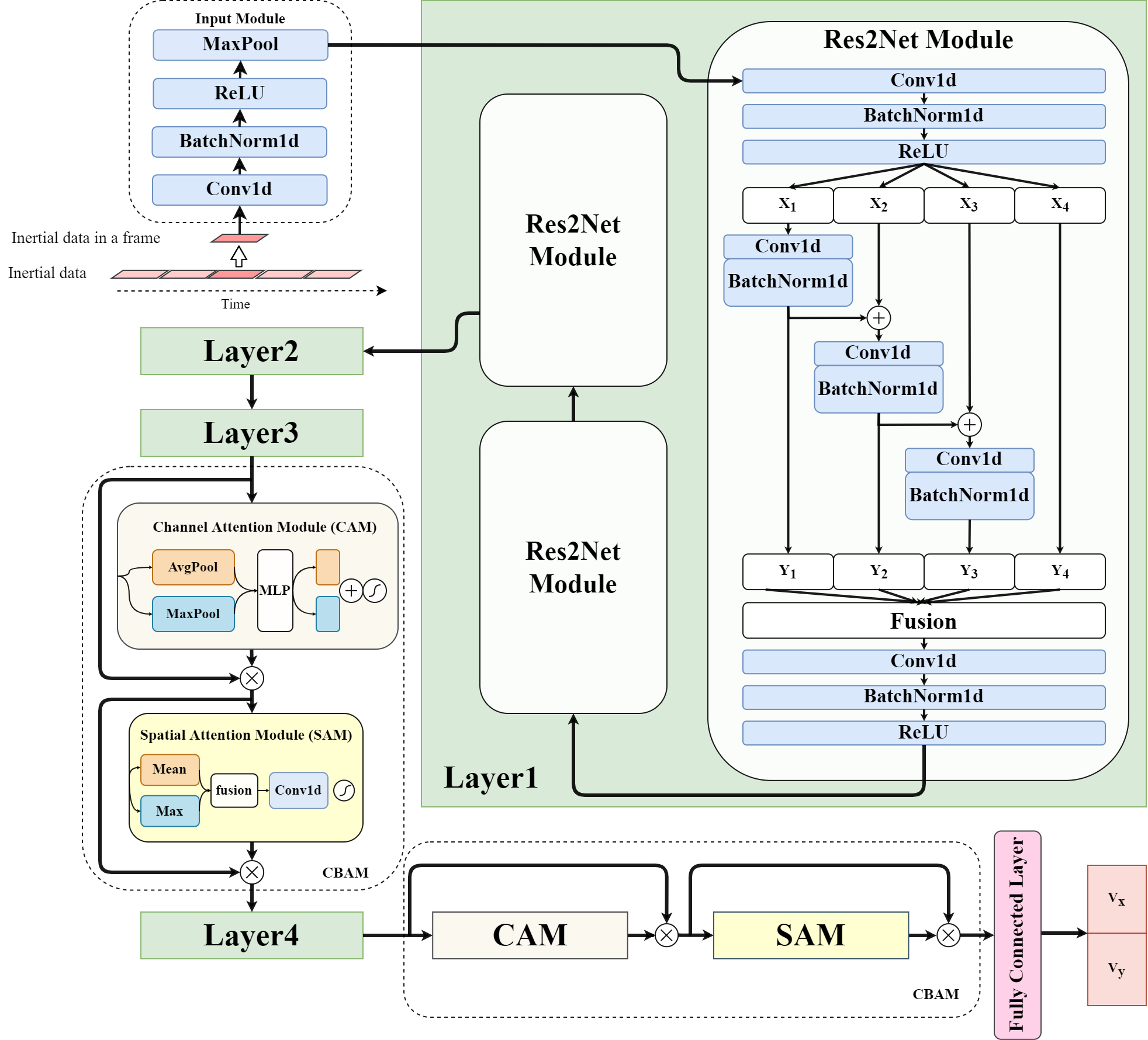}\\
  \caption{Overview of network architecture. A Layer consists of multiple cascaded Res2Net modules. The number of Res2Net modules is different in each layer. The number of Res2Net modules from Layer1 to Layer4 is 3, 4, 6, and 3.}
\end{figure}where $\upsilon_{xi}$ and $\upsilon_{yi}$ are the output velocities of the the network at the $i$th moment, $\hat{\upsilon}_{xi}$ and $\hat{\upsilon}_{yi}$ are the corresponding ground truth velocities, and $n$ is the the total number of data in the training set.


\subsubsection{Res2Net Module}
The Res2Net module structure is a new backbone architecture \cite{Gao2021Res2NetAN} to represent features at multiple scales for numerous vision tasks. It increases the features dimension and reveals a new dimension, namely scale, which increases the range of perceptual fields for each network layer. In this letter, We replace its two-dimensional convolutions with one-dimensional convolutions to handle the continuous time-series learning problem. The detailed structure of Res2Net is shown in Fig. 1.

In the Res2Net module structure, features split into four uniform segments, X1, X2, X3, and X4, will pass through a separate convolution layer. The features of the first segment passing through the convolution layer will add to the features of the second segment before entering the second convolution layer. This process can be considered as another residual operation in the residual block. We end up with four segments of features, Y1, Y2, Y3, and Y4, which are fused with the features through a fusion gate. This module allows better extraction of global and local information.


Every X adds up with the previous features before the convolution operation so that all the information of previous features can be utilized. Therefore the module can get a larger perceptual field. The experiments demonstrate that our idea is practicable.

\subsubsection{Attention Mechanism Module}
The CBAM \cite{Woo2018CBAMCB} is a simple and effective feed-forward convolutional neural network attention module that infers the attention map by channel and spatial dimensions. The attention map is multiplied by the features to get the adaptive feature refinement. This module can enhance the practical features and reduce the noise to help the network learn the relationship between IMU data and velocity more effectively. We adapt two CBAMs to obtain the weighted features. The first CBAM is put between Layer3 and Layer4. The influence of the CBAM placement on the network performance is discussed in the section III-C. The second CBAM is placed before the finial FC layer to further optimize the processed weighted features. 

\subsection{Position Estimation}
The velocity estimation network described in the previous section returns the velocity at each moment, and the displacement can be obtained by integrating the velocity. In the case of the known initial position $(p_{x_{0}}, p_{y_{0}})$, the global position can be written as:
\begin{equation}
\left\{\begin{array}{l}
p_{x_{i}}=p_{x_{0}}+\sum_{1}^{i} \upsilon_{x_{i}} \Delta t \quad 0<i \leq n \\
p_{y_{i}}=p_{y_{0}}+\sum_{1}^{i} \upsilon_{y_{i}} \Delta t \quad 0<i \leq n
\end{array}\right.
\end{equation}
where $\upsilon_{x_{i}}$ and $\upsilon_{y_{i}}$ are the velocities at the moment $i$ of the velocity estimation network regression. $n$ is the total time of the trajectory, and $\Delta t$ is the sample time interval.

\section{Experiments}

\subsection{Dataset}
We evaluate our network on OXIOD \cite{Chen2018OxIODTD} and RoNIN \cite{Yan2020RoNINRN} datasets which are described as follows. 

\textbf{OXIOD}: An inertial odometry dataset with different device placement positions (handheld, in the pocket, in the handbag, and on a trolley). We select three different device placements in pedestrian tracking to evaluate our work.

\textbf{RoNIN}: Another inertial odometry dataset with more data, more human subjects, and more realistic device placement attitudes than OXIOD. For security reasons, the dataset is only published for 50$\%$ of the data, so we use 50$\%$ of the data to evaluate our work. We divide 80$\%$ of the subjects' dataset into training, validating, and testing subsets, and the test set in this part is the seen subjects test set.  The remaining 20$\%$ are used to test the generalization ability of the model for unknown subjects, the test set in this part is the unseen subjects test set.
\subsection{Metrics Definitions}
In order to evaluate the performance of our framework, we apply the following metrics for data of length n.
\subsubsection{The Absolute Translation Error (ATE)}
ATE defines as the root mean square error between the position prediction of the whole trajectory and the ground truth trajectory.
\begin{equation}
    ATE = \sqrt {\frac{1}{n}\sum_{i=1}^{n}\Vert p_i-\hat{p_i}\Vert^2 }
\end{equation}
\subsubsection{The Relative Translation Error (RTE)}
RTE defines as the root mean square error between the predicted difference of the trajectory and the difference between the ground truth trajectory over a fixed time interval. We set 1 min in our evaluation. The RTE serves to calculate the difference in the amount of position shift and is suitable for estimating the drift of the system.
\begin{equation}
    RTE = \sqrt {\frac{1}{n}\sum_{i=1}^{n}\Vert p_{i+\Delta t}-{p_i} - (\hat p_{i+\Delta t}-\hat{p_i})\Vert^2 }
\end{equation}
\subsection{Performance Evaluation}
\subsubsection{Model Training}
We implement the structure of our model in Pytorch \cite{Paszke2017AutomaticDI} and train it using the Adam optimizer \cite{Kingma2015AdamAM} on an Nvidia RTX 3070 GPU with 8GB memory. We use the initial learning rate of 0.001 for training with a batch size of 128, and the learning rate decreases by 0.1 as the metrics on the validation set do not decrease for ten epochs, and the dropout layer is adopted to avoid network overfitting effectively.

\subsubsection{Baselines}
Our proposed network is compared with PDR \cite{Tian2015AnEP}, IONet \cite{Chen2018IONetLT}, and RoNIN \cite{Yan2020RoNINRN} on the above mentioned datasets.

\textbf{PDR}: We utilize the step-length and heading system to update positions with a distance of 0.67 m per step.

\textbf{IONet}: IONet regresses step-length and heading through the network and infers the position based on the regression result. We implement the code locally since the code is not publicly available. We observe that step and heading regressions are ineffective, which is related to the parameter balancing the loss function of the two tasks. For the OXIOD dataset, we train a model for each device placement pose, and for the RoNIN dataset, we train a unified model.

\textbf{RoNIN}: We use the official implementation. For the OXIOD dataset, we train separate models for different placement poses. Because only 50$\%$ data from the RoNIN dataset is now publicly available, we retrain all RoNIN models on this dataset to ensure a fair comparison.

\begin{table}[t]
\begin{center}
\caption{Performance of different CBAM placement. The top three results are highlighted in {\color[HTML]{FE0000}red}, {\color[HTML]{009901}green} and {\color[HTML]{3531FF}blue} colors per column.}
\begin{tabular}{|c|cccc|}
\hline
\multicolumn{1}{|l|}{} & \multicolumn{4}{c|}{RoNIN Dataset} \\ \hline
Test subjects & \multicolumn{2}{c|}{seen} & \multicolumn{2}{c|}{unseen} \\ \hline
Metric & \multicolumn{1}{l|}{ATE (m)} & \multicolumn{1}{l|}{RTE (m)} & \multicolumn{1}{l|}{ATE (m)} & \multicolumn{1}{l|}{RTE (m)} \\ \hline
p1 & \multicolumn{1}{c|}{{\color[HTML]{FE0000} 3.44}} & \multicolumn{1}{c|}{{\color[HTML]{3531FF} 2.57}} & \multicolumn{1}{c|}{{\color[HTML]{3531FF} 5.49}} & {\color[HTML]{3531FF} 4.47} \\ \hline
p2 & \multicolumn{1}{c|}{{\color[HTML]{009901} 3.46}} & \multicolumn{1}{c|}{2.59} & \multicolumn{1}{c|}{5.65} & 4.5 \\ \hline
p3 & \multicolumn{1}{c|}{3.53} & \multicolumn{1}{c|}{{\color[HTML]{009901} 2.54}} & \multicolumn{1}{c|}{{\color[HTML]{FE0000} 5.35}} & {\color[HTML]{FE0000} 4.39} \\ \hline
p4 & \multicolumn{1}{c|}{{\color[HTML]{3531FF} 3.47}} & \multicolumn{1}{c|}{{\color[HTML]{FE0000} 2.53}} & \multicolumn{1}{c|}{{\color[HTML]{009901} 5.42}} & {\color[HTML]{009901} 4.43} \\ \hline
\end{tabular}
\end{center}
\end{table}

\begin{table}[t]
\begin{center}
\caption{Positioning performance. We compare four inertial positioning methods: PDR, IONet, Ronin (3 variants) and our method on two public datasets (OXIOD and RoNIN). The top two results are highlighted in {\color[HTML]{FE0000}red} and {\color[HTML]{009901}green} colors per column, respectively.}
\setlength\tabcolsep{4pt}
\begin{tabular}{|c|cc|c|cc|}
\hline
 & \multicolumn{2}{c|}{} &  & \multicolumn{2}{c|}{RoNIN Dataset} \\ \cline{1-3} \cline{5-6} 
Metric & \multicolumn{2}{c|}{Method} & \multirow{-2}{*}{OXIOD Dataset} & \multicolumn{1}{c|}{seen} & unseen \\ \hline
 & \multicolumn{2}{c|}{PDR} & 7.06 & \multicolumn{1}{c|}{25.32} & 23.49 \\ \cline{2-6} 
 & \multicolumn{2}{c|}{IONet} & 2.45 & \multicolumn{1}{c|}{29.24} & 25.07 \\ \cline{2-6} 
 & \multicolumn{2}{c|}{RoNIN-ResNet} & {\color[HTML]{FE0000} 1.08} & \multicolumn{1}{c|}{{\color[HTML]{009901} 3.69}} & {\color[HTML]{009901} 5.74} \\ \cline{2-6} 
 & \multicolumn{2}{c|}{RoNIN-LSTM} & 2.86 & \multicolumn{1}{c|}{5.06} & 6.83 \\ \cline{2-6} 
 & \multicolumn{2}{c|}{RoNIN-TCN} & {\color[HTML]{000000} 4.04} & \multicolumn{1}{c|}{4.55} & 6.28 \\ \cline{2-6} 
\multirow{-6}{*}{ATE (m)} & \multicolumn{2}{c|}{Ours} & {\color[HTML]{009901} 1.18} & \multicolumn{1}{c|}{{\color[HTML]{FE0000} 3.47}} & {\color[HTML]{FE0000} 5.42} \\ \hline
 & \multicolumn{2}{c|}{PDR} & 3.17 & \multicolumn{1}{c|}{22.56} & 23.07 \\ \cline{2-6} 
 & \multicolumn{2}{c|}{IONet} & 1.32 & \multicolumn{1}{c|}{23.29} & 22.33 \\ \cline{2-6} 
 & \multicolumn{2}{c|}{RoNIN-ResNet} & {\color[HTML]{009901} 1.03} & \multicolumn{1}{c|}{{\color[HTML]{009901} 2.69}} & 4.51 \\ \cline{2-6} 
 & \multicolumn{2}{c|}{RoNIN-LSTM} & 2.68 & \multicolumn{1}{c|}{2.73} & 4.55 \\ \cline{2-6} 
 & \multicolumn{2}{c|}{RoNIN-TCN} & 3.16 & \multicolumn{1}{c|}{2.87} & {\color[HTML]{FE0000} 4.41} \\ \cline{2-6} 
\multirow{-6}{*}{RTE (m)} & \multicolumn{2}{c|}{Ours} & {\color[HTML]{FE0000} 0.97} & \multicolumn{1}{c|}{{\color[HTML]{FE0000} 2.53}} & {\color[HTML]{009901} 4.43} \\ \hline
 & \multicolumn{1}{c|}{} & RoNIN-ResNet & -9.3\% & \multicolumn{1}{c|}{6.0\%} & 5.6\% \\ \cline{3-6} 
 & \multicolumn{1}{c|}{} & RoNIN-LSTM & 58.7\% & \multicolumn{1}{c|}{31.4\%} & 20.6\% \\ \cline{3-6} 
 & \multicolumn{1}{c|}{\multirow{-3}{*}{ATE}} & RoNIN-TCN & 70.8\% & \multicolumn{1}{c|}{23.7\%} & 13.7\% \\ \cline{2-6} 
 & \multicolumn{1}{c|}{} & RoNIN-ResNet & 5.8\% & \multicolumn{1}{c|}{5.9\%} & 1.8\% \\ \cline{3-6} 
 & \multicolumn{1}{c|}{} & RoNIN-LSTM & 46.5\% & \multicolumn{1}{c|}{7.3\%} & 2.6\% \\ \cline{3-6} 
\multirow{-6}{*}{\begin{tabular}[c]{@{}c@{}}Our \\ improvement\\ over\\ RoNIN\end{tabular}} & \multicolumn{1}{c|}{\multirow{-3}{*}{RTE}} & RoNIN-TCN & 69.3\% & \multicolumn{1}{c|}{11.8\%} & -0.5\% \\ \hline
\end{tabular}
\end{center}
\end{table}
\subsubsection{Results and Analysis}

Table I discusses the effect of the first CBAM placements on positioning performance. p1, p2, p3, and p4 represent different placements of the module. p1 is in front of Layer1, p2 between Layer1 and Layer2, p3 from Layer2 and Layer3, and p4 between Layer3 and Layer4. According to the results, the module performs best at p4, so we place it between Layer3 and Layer4.


Table II shows our main results. It shows that our method achieves the best ATE and RTE on most data sets. However, our method is the second best in two cases. RoNIN-ResNet can get a smaller ATE than our method on the OXIOD dataset. RoNIN-TCN achieves the best RTE on the RoNIN unseen dataset. Our method improves the ATE on the RoNIN seen test dataset by 6$\%$, 21.78$\%$ and 37.46$\%$ over RoNIN-ResNet, RoNIN-LSTM, and RoNIN-TCN, respectively. The corresponding results for the RoNIN unseen test dataset are 5.6$\%$, 20.6$\%$, and 13.7$\%$. In summary, considering the total performance on both datasets, we believe our method is the best performing model among all discussed models. The key to our method that makes the result better may be the combination of the previous weighted features and the current weighted features in a larger perceptual field.

\begin{figure}[t]
\includegraphics[width=\linewidth]{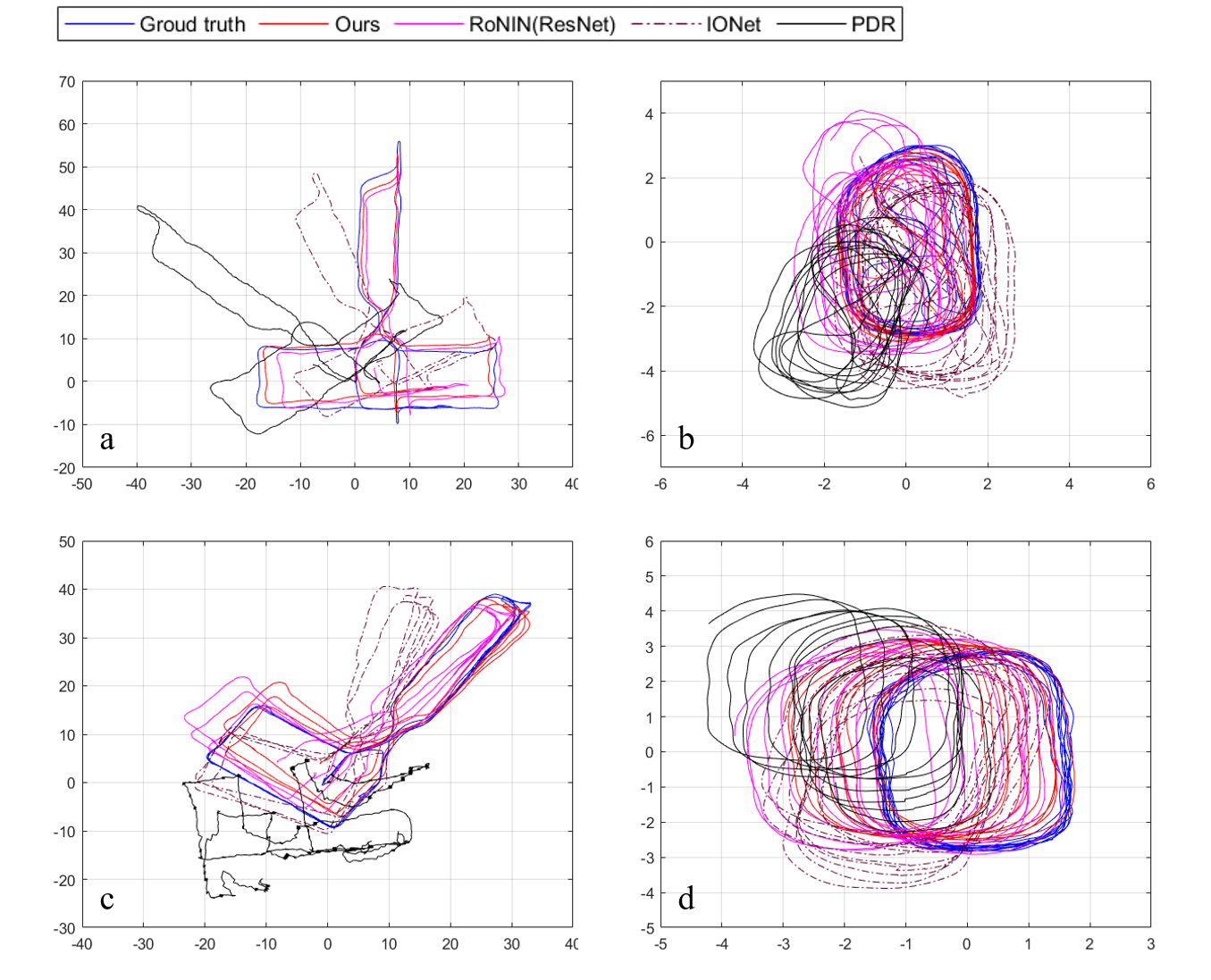}
\caption{Selected trajectories. We select two examples from each dataset and visualize their trajectories, (a) and (c) are from the RoNIN dataset, while (b) and (d) are from the OXIOD dataset.}
\end{figure}

\begin{figure}[t]
\includegraphics[width=\linewidth]{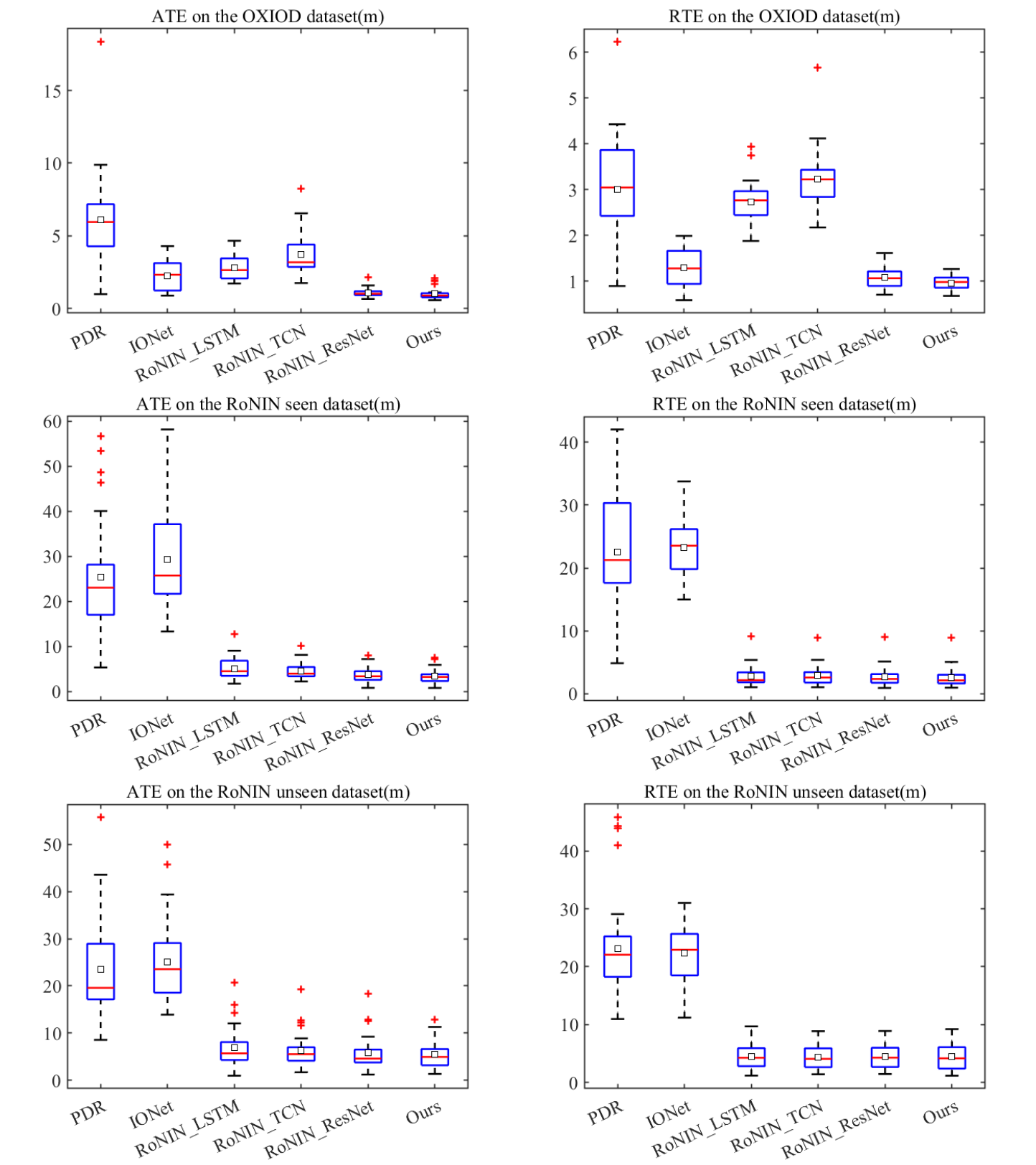}
\caption{Box plots of the two metrics for the OXIOD dataset and the RoNIN dataset. The first row is from the oxiod dataset. The second row represents the seen set of the RONIN dataset, and the third row represents the unseen set of the RONIN dataset. The squares in all sub-figures represent the mean.}
\end{figure}

Fig. 2 shows the visualization of reconstructed the trajectory, demonstrating our method's effectiveness. Fig. 3 shows the box plots of two evaluation metrics for all seen test sets and unseen test sets on the OXIOD and RoNIN datasets. On the OXIOD dataset, our method outperforms the comparing methods, and the data distribution is more concentrated than those of other methods. However, the ATE has more outlier points, which may be related to the error of the device orientation provided by the dataset sequence. On the seen test set of the RONIN dataset, our method not only has the smallest ATE and RTE but also has minor data fluctuation and concentrated data distribution. The effect of our method is feasible in the unseen test set. Although the data distribution is similar to other methods, the outlier points are minimal, indicating our method's stability. The results of the seen data sets show that our method achieves good results for the localization of known human subjects. In contrast, the localization of unknown human subjects needs stronger robustness of data-driven deep learning methods because of the different testing data distribution (such as walking habits) from the training data, which will be investigated in our future work.

\section{Conclusion}

In this letter, we propose a deep learning-based inertial odometry that introduces Res2Net modules and two CBAMs to the ResNet50 framework to estimate the position by regressing the two-dimensional velocity. Experimental results show that our method improves the ATE on the RoNIN dataset by 76$\%$-86$\%$ over the traditional PDR algorithm and improves the ATE by 6$\%$, 31.4$\%$, and 23.7$\%$ compared to the state-of-the-art deep learning-based inertial odometry (RoNIN-ResNet, RoNIN-LSTM, RoNIN-TCN), respectively.



\bibliographystyle{IEEEtran}
\bibliography{cited}

\begin{thebibliography}{10}
\providecommand{\url}[1]{#1}
\csname url@samestyle\endcsname
\providecommand{\newblock}{\relax}
\providecommand{\bibinfo}[2]{#2}
\providecommand{\BIBentrySTDinterwordspacing}{\spaceskip=0pt\relax}
\providecommand{\BIBentryALTinterwordstretchfactor}{4}
\providecommand{\BIBentryALTinterwordspacing}{\spaceskip=\fontdimen2\font plus
\BIBentryALTinterwordstretchfactor\fontdimen3\font minus
  \fontdimen4\font\relax}
\providecommand{\BIBforeignlanguage}[2]{{%
\expandafter\ifx\csname l@#1\endcsname\relax
\typeout{** WARNING: IEEEtran.bst: No hyphenation pattern has been}%
\typeout{** loaded for the language `#1'. Using the pattern for}%
\typeout{** the default language instead.}%
\else
\language=\csname l@#1\endcsname
\fi
#2}}
\providecommand{\BIBdecl}{\relax}
\BIBdecl

\bibitem{Yang2012LocatingIF}
Z.~Yang, C.~Wu, and Y.~Liu, ``Locating in fingerprint space: wireless indoor
  localization with little human intervention,'' in \emph{Mobicom '12}, 2012.

\bibitem{Au2013IndoorTA}
W.~S.~A. Au, C.~Feng, S.~Valaee, S.~Reyes, S.~Sorour, S.~N. Markowitz, D.~Gold,
  K.~Gordon, and M.~Eizenman, ``Indoor tracking and navigation using received
  signal strength and compressive sensing on a mobile device,'' \emph{IEEE
  Transactions on Mobile Computing}, vol.~12, pp. 2050--2062, 2013.

\bibitem{Harle2013ASO}
R.~K. Harle, ``A survey of indoor inertial positioning systems for
  pedestrians,'' \emph{IEEE Communications Surveys \& Tutorials}, vol.~15, pp.
  1281--1293, 2013.

\bibitem{Foxlin2005PedestrianTW}
E.~Foxlin, ``Pedestrian tracking with shoe-mounted inertial sensors,''
  \emph{IEEE Computer Graphics and Applications}, vol.~25, pp. 38--46, 2005.

\bibitem{Tian2016AMD}
Q.~Tian, Z.~A. Salcic, K.~I.-K. Wang, and Y.~Pan, ``A multi-mode dead reckoning
  system for pedestrian tracking using smartphones,'' \emph{IEEE Sensors
  Journal}, vol.~16, pp. 2079--2093, 2016.

\bibitem{Leutenegger2015KeyframebasedVO}
S.~Leutenegger, S.~Lynen, M.~Bosse, R.~Y. Siegwart, and P.~T. Furgale,
  ``Keyframe-based visual–inertial odometry using nonlinear optimization,''
  \emph{The International Journal of Robotics Research}, vol.~34, pp. 314 --
  334, 2015.

\bibitem{Brajdic2013WalkDA}
A.~Brajdic and R.~K. Harle, ``Walk detection and step counting on unconstrained
  smartphones,'' \emph{Proceedings of the 2013 ACM international joint
  conference on Pervasive and ubiquitous computing}, 2013.

\bibitem{Yan2018RIDIRI}
H.~Yan, Q.~Shan, and Y.~Furukawa, ``Ridi: Robust imu double integration,'' in
  \emph{ECCV}, 2018.

\bibitem{Chen2018IONetLT}
C.~Chen, C.~X. Lu, A.~Markham, and A.~Trigoni, ``Ionet: Learning to cure the
  curse of drift in inertial odometry,'' in \emph{AAAI}, 2018.

\bibitem{Chen2021DeepNN}
C.~Chen, X.~Lu, J.~Wahlstrom, A.~Markham, and N.~Trigoni, ``Deep neural network
  based inertial odometry using low-cost inertial measurement units,''
  \emph{IEEE Transactions on Mobile Computing}, vol.~20, pp. 1351--1364, 2021.

\bibitem{Yan2020RoNINRN}
H.~Yan, S.~Herath, and Y.~Furukawa, ``Ronin: Robust neural inertial navigation
  in the wild: Benchmark, evaluations, \& new methods,'' \emph{2020 IEEE
  International Conference on Robotics and Automation (ICRA)}, pp. 3146--3152,
  2020.

\bibitem{He2016DeepRL}
K.~He, X.~Zhang, S.~Ren, and J.~Sun, ``Deep residual learning for image
  recognition,'' \emph{2016 IEEE Conference on Computer Vision and Pattern
  Recognition (CVPR)}, pp. 770--778, 2016.

\bibitem{Gao2021Res2NetAN}
S.~Gao, M.-M. Cheng, K.~Zhao, X.~Zhang, M.-H. Yang, and P.~H.~S. Torr,
  ``Res2net: A new multi-scale backbone architecture,'' \emph{IEEE Transactions
  on Pattern Analysis and Machine Intelligence}, vol.~43, pp. 652--662, 2021.

\bibitem{Woo2018CBAMCB}
S.~Woo, J.~Park, J.-Y. Lee, and I.-S. Kweon, ``Cbam: Convolutional block
  attention module,'' in \emph{ECCV}, 2018.

\bibitem{Chen2018OxIODTD}
C.~Chen, P.~Zhao, C.~X. Lu, W.~Wang, A.~Markham, and A.~Trigoni, ``Oxiod: The
  dataset for deep inertial odometry,'' \emph{ArXiv}, vol. abs/1809.07491,
  2018.

\bibitem{Paszke2017AutomaticDI}
A.~Paszke, S.~Gross, S.~Chintala, G.~Chanan, E.~Yang, Z.~DeVito, Z.~Lin,
  A.~Desmaison, L.~Antiga, and A.~Lerer, ``Automatic differentiation in
  pytorch,'' 2017.

\bibitem{Kingma2015AdamAM}
D.~P. Kingma and J.~Ba, ``Adam: A method for stochastic optimization,''
  \emph{CoRR}, vol. abs/1412.6980, 2015.

\bibitem{Tian2015AnEP}
Q.~Tian, Z.~A. Salcic, K.~I.-K. Wang, and Y.~Pan, ``An enhanced pedestrian dead
  reckoning approach for pedestrian tracking using smartphones,'' \emph{2015
  IEEE Tenth International Conference on Intelligent Sensors, Sensor Networks
  and Information Processing (ISSNIP)}, pp. 1--6, 2015.

\end{thebibliography}

\end{document}